\documentclass[runningheads]{llncs}
\usepackage[T1]{fontenc}
\usepackage{graphicx}
\usepackage{subcaption}
\begin{document}
\title{A Novel Neural Network Training Method for Autonomous Driving Using Semi-Pseudo-Labels and 3D Data Augmentations}
%
%
\author{Tamás Matuszka \and
Dániel Kozma}
\authorrunning{T. Matuszka, D. Kozma}
%
\institute{aiMotive, Budapest, Hungary \\
\email{\{tamas.matuszka,daniel.kozma\}@aimotive.com}\\
\url{https://aimotive.com/}}
\maketitle              
\begin{abstract}
Training neural networks to perform 3D object detection for autonomous driving requires a large amount of diverse annotated data. However, obtaining training data with sufficient quality and quantity is expensive and sometimes impossible due to human and sensor constraints. Therefore, a novel solution is needed for extending current training methods to overcome this limitation and enable accurate 3D object detection. Our solution for the above-mentioned problem combines semi-pseudo-labeling and novel 3D augmentations. For demonstrating the applicability of the proposed method, we have designed a convolutional neural network for 3D object detection which can significantly increase the detection range in comparison with the training data distribution.

\end{abstract}
\section{Introduction}
\label{sec:intro}
Object detection is a crucial part of an autonomous driving software since increasingly complex layers are built on the top of the perception system which itself relies on fast and accurate obstacle detections. Object detection is typically performed by convolutional neural networks which are trained by means of supervised learning. Supervised learning is a method where a model is fed with input data and its main objective is to learn a function that maps the input data to the corresponding output. Since convolutional neural networks, the best models for visual domain (except large-data regime where visual transformers \cite{Kolesnikov2021} excel), are heavily overparameterized, a large amount of annotated data is needed for learning the mapping function. Therefore, a substantial amount of manual effort is required to annotate data with sufficient quality and quantity, which is an expensive and error-prone method. In addition, obtaining precise ground truth data is sometimes impossible due to human or sensor constraints. For example, the detection range of LiDARs limits the annotation of distant objects which presence must be known by an autonomous driving system due to fast position and location change on a highway. Radars could overcome this kind of limitation, but they have a more reduced field of view and only a subset of the interesting object categories are detectable with it. A human limitation is, for example, the inability to accurately estimate the distance of objects in 3D from 2D images without any 3D clue e.g. point clouds collected by a LiDAR or radar detections. Consequently, a novel solution is needed for extending current training methods to overcome this limitation and enable accurate 3D object detection. 

Several approaches have been developed to facilitate neural network training. One of the most popular solutions is transfer learning where a neural network is trained on a particular dataset, such as ImageNet \cite{FeiFei2009} and then fine-tuned on another dataset (e.g., a model trained to recognize cars can be trained to classify trucks using transfer learning). Self-supervised learning which utilizes unlabeled data to train a model performing proxy tasks and then fine-tunes it on a downstream task in a supervised manner has resulted in breakthroughs in language modeling. Pseudo-labeling is a simple solution that uses the same model’s predictions as true labels during the training. However, none of these solutions helps the model to produce predictions that are not part of the training distribution. 

The main motivation of this work is to develop a training method which massively overcomes the limitations of the training dataset and so extends the prediction capabilities of a neural network. To summarize, this paper makes the following three main contributions:
\begin{itemize}
  \item We introduced semi-pseudo-labeling (SPL) as a method where pseudo-labels are generated by a neural network trained on a simpler task and utilized during the training of another network performing a more complex task.
  \item We extended several conventional 2D data augmentation methods to work in 3D. 
  \item We described a training method for allowing neural networks to predict certain characteristics from out of training distribution using semi-pseudo-labeling and 3D data augmentations. 
\end{itemize}

\section{Related Work}
\label{sec:relatedwork}
The concept of pseudo-labeling has been introduced by Lee in \cite{Lee_pseudo-label} as a simple and efficient self-supervised method for deep neural networks. The main idea of pseudo-labeling is to consider the predictions of a trained model as ground truth. Unlabeled data, which is typically easy to obtain, can be annotated using the trained model’s prediction. Then, the same model is retrained on the labeled and pseudo-labeled data simultaneously. Our proposed method is based on this concept but there is a fundamental difference between the solutions. Pseudo-labeling generates labels for the same task using the same model while our semi-pseudo-labeling method utilizes pseudo-labels generated by a different model for a more complex task. In \cite{Chen2018}, Chen generated pseudo-labels for object detection on dynamic vision sensor using a convolutional neural network trained on an active pixel sensor. The main difference compared to our solution is that pseudo-labels in the paper are used for the same task, namely two-dimensional bounding box detection of cars on different sensor modalities. The solution described in \cite{Yu2020} also uses pseudo-labels for training object detection neural networks. However, the invention described in the patent uses regular pseudo-labeling for training a neural network to perform the same task, namely, 2D object detection as opposed to our solution where the tasks are not the same. In addition, the solution requires the use of region proposal networks which indicates a two-stage network architecture that might not fulfill real-time criterion while our 3D object detection network uses a single stage architecture that utilizes semi-pseudo-labeling during training. Watson et al. used pseudo-labeling in \cite{Watson2021} for generating and augmenting their data labeling method. Their proposed solution has created pseudo-labels for unlabeled data while our method enables us to use annotated data created for a simpler task as pseudo-labels and does not exclusively rely on unlabeled data. 

Transfer learning \cite{Bozinovski2020} is the process when a model is trained for performing a specific task and its knowledge is utilized to solve another (related) problem. Transfer learning involves pretraining (typically on a large-scale dataset) a model and then customizing this model to a given task by either using the trained weights or fine-tuning the model by adding an additional classifier on the top of frozen weights. Transfer learning can be effectively used to transfer knowledge between related problems which has similarities with our proposed method which can be considered as an extended version of transfer learning. However, we utilize semi-pseudo-labels to perform a more complex task (e.g. 3D object detection with 2D pseudo labels) which might not be solvable using regular transfer learning. In addition, our solution enables simultaneous learning of different tasks as opposed to transfer learning. 

Data augmentation \cite{Xu2020} is a standard technique for enhancing training data and preventing overfitting using geometric transformations, color space augmentations, random erasing, mixing images, etc. Most data augmentation techniques operate in image space (2D) \cite{Liu2020}, but recent work started to extend their domain to 3D \cite{Xu2020}, \cite{Liu2020}. To the best of our knowledge, none of these solutions introduced zoom augmentation in 3D using a virtual camera, as our work proposes. The closest solution which tries to solve the limited perception range is described in \cite{Simonelli2020}. The method in the paper proposes to break down the entire image into multiple image patches, each containing at least one entire car and with limited depth variation. During inference, a pyramid-like tiling of images is generated which increases the running time. In addition, the perception range of the approach described in the paper did not exceed 50 meters. 

\section{A Novel Training Method with Semi-Pseudo-Labeling and 3D Augmentations}
\label{sec:method}
Two methods have been developed for enhancing training dataset limitations, namely semi-pseudo-labeling (SPL) and 3D augmentations (zoom and shift). SPL is first introduced as a general, abstract description. Then, the concept and its combination with 3D augmentations is detailed using a concrete example.

\subsection{Semi-Pseudo-Labeling}
The main objective of supervised learning is to define a mapping from input space to the output by learning the value of parameters that minimizes the error of the approximation function \cite{Goodfellow2016}, formally
\begin{equation}
  M(X;\theta)=Y
  \label{eq:sl}
\end{equation}
and
\begin{equation}
  \min_{\theta} L( Y - M(X, \theta) )
  \label{eq:min_loss}
\end{equation}
where $L$ is an arbitrary loss function, $Y$ is the predictable target, $X$ is the input and $M$ is the model parameterized by $\theta$. 

\noindent For training a model using supervised learning, a training set is required: 
\begin{equation}
  D_{GT}=\left\{ \left(x_{1}^{L}, y_{1}^{L}  \right), \ldots , \left(x_{n}^{L}, y_{n}^{L}  \right) \right\}\subseteq R^{d} \times C
  \label{eq:ts}
\end{equation}
where $x_{1}^{L} \in X^{L}$ is the input sample where annotations are available, $y_{1}^{L} \in Y^{L}$ is the ground truth, $R^{d}$ is the $d$ dimensional input space, $C$ is the label space. 

The pseudo-labeling method introduces another dataset where labels for an unlabeled dataset are generated by the trained model $M(X;\theta)$: 
\begin{equation}
  D_{PS}=\left\{ \left(x_{1}^{U}, y_{1}^{PS}  \right), \ldots , \left(x_{m}^{U}, y_{m}^{PS}  \right) \right\}\subseteq R^{d} \times C
  \label{eq:ps_ts}
\end{equation}
where $x_{i}^{U} \in X^{U}$ is the input sample from unlabeled data, $y_{i}^{PS} \in M(X^{U};\theta)$ is the pseudo-label generated by the trained model.

The final model is trained on the union of the annotated and pseudo-labeled datasets. The main objective of semi-pseudo-labeling is to utilize pseudo-labels generated by a model  trained on a simpler task for training another model performing a more complex task. Both simple and complex tasks have annotated training sets for their specific tasks: 
\begin{equation}
    D_{GT}^{S}=\left\{ \left(x_{1}^{SL}, y_{1}^{SL}  \right), \ldots , \left(x_{n}^{SL}, y_{n}^{SL}  \right) \right\}\subseteq R^{d} \times C^{S}
  \label{eq:s_gt}
\end{equation}
\begin{equation}
    D_{GT}^{C}=\left\{ \left(x_{1}^{CL}, y_{1}^{CL}  \right), \ldots , \left(x_{m}^{CL}, y_{m}^{CL}  \right) \right\}\subseteq R^{d} \times C^{C}
  \label{eq:c_gt}
\end{equation}

The main differentiator between regular pseudo-labeling and semi-pseudo-labeling method is that the simple model $M^{S}$ does not generate pseudo-labels on unlabeled data (although it is a viable solution and might be beneficial in some cases). Rather, pseudo-labels are generated using the input data of the complex model $M^{C}$. In this way, the label space of complex model can be extended, as can be seen in 
\ref{eq:sps_gt}: 
\begin{equation}
    D_{SPS}^{C}=\left\{ \left(x_{1}^{CL}, y_{1}^{CL}\cup y_{1}^{SPS}   \right), \ldots , \left(x_{m}^{CL}, y_{m}^{CL} \cup y_{m}^{SPS}  \right) \right\} \subseteq R^{d} \times \left(C^{C} \cup C^{S} \right)
  \label{eq:sps_gt}
\end{equation}
where $x_{i}^{CL} \in X^{CL}$ is the input sample where annotations for the complex task are available, $y_{i}^{CL} \in Y^{CL}$ is the ground truth label for the complex task, $y_{i}^{SPS} \in M^{S}(X^{CL};\theta^{S})$ is the semi-pseudo-label generated by the simple model on the complex task's input. The final model $M^{C}(X^{CL};\theta^{C})$ is trained on the semi-pseudo-labeled $D_{SPS}^{C}$ dataset.

\subsection{3D Augmentations}
\subsubsection{Issues with Vanilla Zoom Augmentation}
\label{sec:vanilla}
\begin{figure}[t]
  \centering
  \includegraphics[width=0.6255\linewidth]{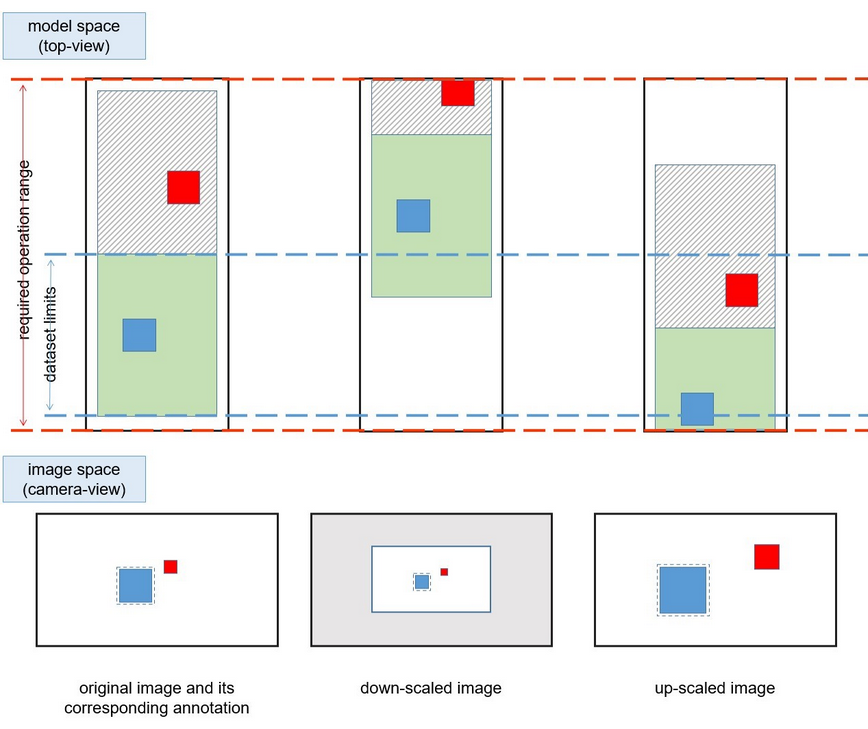}
   \caption{Effects of zoom augmentation on image and model space data.}
   \label{fig:aug}
\end{figure}

For training a network predicting 3D attributes of dynamic objects, accurate 3D position, size, and orientation data is required in model space. The principal problem to solve for overcoming the dataset limitations is handling image-visible, non-annotated objects (red squares on Fig. \ref{fig:aug}) within the required operational distance range (enclosed by red dashed lines in Fig. \ref{fig:aug}). 

Gray dashed area presents the non-annotated region and red square is non-annotated object, while green area is the annotated region and blue square is an annotated object. Red dashed horizontal lines represent the required operational domain, in which the developed algorithm has to detect all the objects, while blue dashed horizontal lines are the distance limits of the annotated data. The three columns represent the three options during zoom augmentation. First case (left) is the non-augmented, original version. Second (center) case when input image is downscaled, mimicking farther objects in image space, so the corresponding ground truth in the model space should be adjusted consistently. The third case (right), when input image is upscaled, bringing the objects closer to the camera. The figure highlights the inconsistency between applying various zoom levels, as annotated regions on the transformed cases (second, third) overlap with the original non-annotated region. 

Figure \ref{fig:aug} presents the inconsistencies of applying vanilla zoom augmentation. The green area represents the region where all image-visible objects are annotated, while the gray dashed one is, where our annotation is imperfect and contains false negatives. When applying vanilla zoom augmentation technique to extend the operational domain of the developed algorithm, discrepancies may arise, i.e., when the zoom augmented dataset contains original and down-scaled images (case \#1 and \#2) on Fig. \ref{fig:aug}, it can be seen ground truth frames contradict each other. In case \#2 it is required to detect objects beyond the original ground truth limit (upper blue dashed line), while in case \#1 it cannot be utilized in the loss function, since there is no available information on even the existence of the red object. To overcome the limitations and make zoom augmentation viable in our case, additional information is required to fill missing data, i.e., non-annotated objects at least in image space. Lacking data could be refill with human supervision but this is infeasible since it is not scalable. Pseudo-labeling is a promising solution. However, in our case, the whole 3D information could not be recovered but the existence of 2D information is sufficient to overcome the above-mentioned issues. Therefore, 3D zoom augmentation becomes a viable solution for widening the limits of the dataset and extending the operational domain of the developed detection algorithm. A pretrained, state-of-the-art 2D bounding box network can be used to detect all image-visible objects.

\subsubsection{Improving over Existing Augmentations}
Most 2D data augmentations are easy to generalize to three-dimension. However, zooming is not trivial since changes in the image scale modify the position and egocentric orientation of the annotations in 3D space too. A 3D zoom augmentation using a virtual camera has been developed to resolve this issue. The method consists of two main steps. The first is to either zoom in or zoom out of the image. In this way, it can be emulated that an object moves either closer or farther to the camera. The second step is to modify the camera matrix to follow the 2D transformations and to keep 3D annotations intact. This can be performed by linear transformations and a virtual camera that adjusts its principal point and focal length considering the original camera matrix and the 2D scaling transformation. Changing camera intrinsic parameters mimics the change of the egocentric orientation of the given object, but its apparent orientation, which is the regressed parameter during the training, remains the same. 

The 3D zoom augmentation can be implemented as follows. As a first step, a scaling factor between an empirically chosen lower and upper bound is randomly drawn. If the lower and upper bound is smaller than one, a zoom-out operation is performed. If the lower and upper bound are greater than one, a zoom-in operation is performed. If the lower bound is less than one and the upper bound is greater than one, either a zoom-in or zoom-out is performed. The 2D part of zoom works as in the traditional case when one zooms in / out to the image using the above-mentioned scaling factor (in the case of zooming out, the image is padded with zeros for having the original image size). Then, the camera matrix corresponding to the image can be adjusted by scaling the focal length components with the randomly drawn scaling factor. If the 2D image is shifted beside the zoom operation, the camera matrix can be adjusted by shifting the principal point components. Therefore, the augmentations for a corresponding image and 3D labels are performed in a consistent manner. 

Applying random shift of the image enforces the decoupling of image position and object distance. Due to this augmentation, the detection system can prevent overfitting to a specific camera intrinsics.

\subsection{An Example of Training with Semi-Pseudo-Labeling and 3D Augmentations}
\begin{figure}[t]
  \centering
  \includegraphics[width=0.43\linewidth]{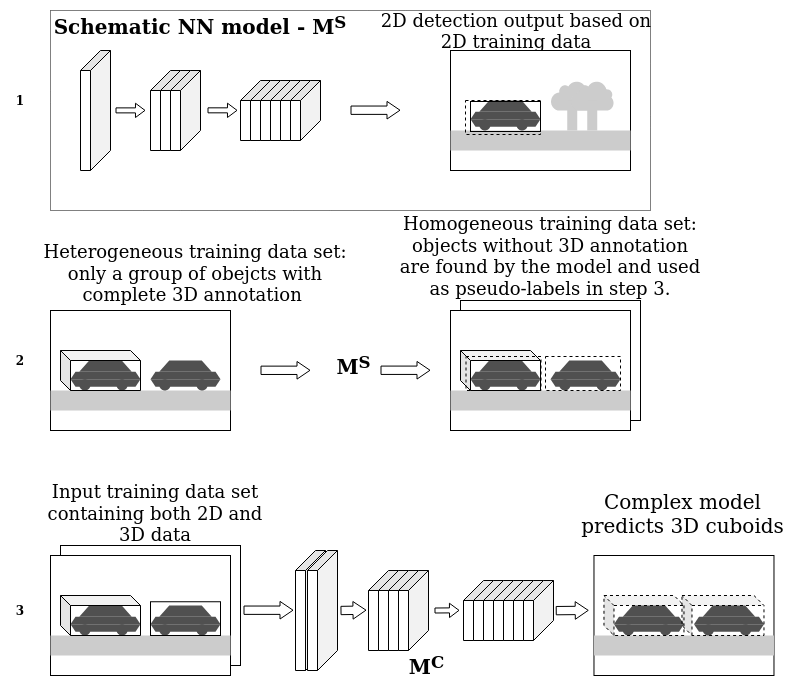}
   \caption{SPL applied in 3D object detection using 2D detection as the simple task.}
   \label{fig:method}
\end{figure}

The semi-pseudo-labeling method combined with 3D augmentations was used for training a 3D object detection neural network to perform predictions that are out of the training data distribution. Figure \ref{fig:method} describes the steps of the utilization of the SPL method. The requirement was to extend the detection range of an autonomous driving system to 200 meters while the distance range of annotated data did not exceed 120 meters. In addition, some detectable classes were missing from the training data.

The FCOS \cite{Tian2019} 2D bounding box detector has been chosen as the simple model $M^{S}(X^{SL};\theta^{S})$ where the input space $X^{SL}$ contains HD resolution stereo image pairs and label space $C^S$ consists of $\left( x,y,w,h,o,c_{1}, \ldots, c_{n} \right)$ tuples, where $x$ is the x coordinate of the bounding box center in image space, $y$ is the y coordinate of the bounding box center in image space, $w$ is the width of the bounding box in image space, $h$ is the height of the bounding box in image space, $o$ is the objectness score, $c_{i}$ is the probability that the object belongs to the $i$-th category. 

The model $M^{S}$ has performed 2D object detections on the 3D annotated dataset which in our case is the same as $X^{SL}$, i.e., HD images. The resulting 2D detections found distant objects which are not annotated by 3D bounding boxes and added as semi-pseudo-labels. Finally, the 3D object detector was trained on the combination of 3D annotated data and semi-pseudo-labeled 2D bounding boxes. The label space of the 3D detector is $\left( x,y,w,h,o,c_{1}, \ldots, c_{n},P,D,O \right)$ tuples, where $x$ is the x coordinate of the bounding box center in image space, $y$ is the y coordinate of the bounding box center in image space, $w$ is the width of the bounding box in image space, $h$ is the height of the bounding box in image space, $o$ is the objectness score, $c_{i}$ is the probability that the object belongs to the $i$-th category, $P$ is a three-dimensional vector of the center point of 3D bounding box in model space, $D$ is a three-dimensional vector containing the dimensions (width, height, length) of the 3D bounding box, $O$ is a four-dimensional vector of the orientation of the 3D bounding box represented as a quaternion. 

\label{sec:example}

A deduplication algorithm is required to avoid double annotations that are included both in the semi-pseudo-labeled dataset and in the 3D annotated ground truth. This post-processing step can be executed by examining the ratio of the intersection over union (IoU) of the semi-pseudo-labeled annotations and 2D projection of 3D bounding boxes. If the ratio exceeds a threshold, the pseudo-labeled annotation should be filtered out.  
\subsubsection{Baseline Neural Network and Training}
\label{sec:nn}
We have developed a simple single-stage object detector based on the YOLOv3 \cite{Redmon2018} convolutional neural network architecture which has utilized our semi-pseudo-labeling method and 3D data augmentations during its training. The simple architecture was a design choice taking into account two reasons. First, the model has to be lightweight in order to be able to run real-time in a computationally constrained environment (i.e. in a self-driving car). Second, a simple model facilitates the benchmarking of the effects of the proposed methods. As the first step of the training, the input image is fed to an Inception-ResNet \cite{Szegedy2017} backbone. Then, the resulting embedding is passed to a Feature Pyramid Network \cite{Lin2017}. The head is adapted from the YOLOv3 paper and is extended with channels that are responsible for predicting 3D characteristics mentioned above. 

The neural network has been trained using multitask learning \cite{Caruana1997}, 2D (using previously generated semi-pseudo-labels) and 3D detection are learned in a parallel manner. Instead of directly learning the 3D center point of the cuboid, the network was designed to predict the 2D projection of the center of a 3D cuboid. The center point of the 3D bounding box can later be reconstructed from the depth and its 2D projection. Finally, the dimension prediction part of the network uses priors (i.e., precomputed category averages), and only the differences from these statistics are predicted instead of directly regressing the dimensions. This approach was inspired by the YOLOv3 architecture which uses a similar solution for bounding box width and height formulation. 3D zoom augmentation was performed during the training where the lower and upper bound of scaling factor hyperparameters were set to 0.5 and 2.0, respectively.

\subsubsection{Loss Functions}
The label space of semi-pseudo-labels is more restricted than the 3D label space since SPLs (i.e., 2D detections) do not contain 3D characteristics. The ground truth was extended with a boolean flag that indicates whether the annotated object is a semi-pseudo-label or not. This value was used in the loss function to mask out 3D loss terms in the case of semi-pseudo-labels to not penalize the weights corresponding to 3D properties during backpropagation when no ground truth is known. Due to this solution and the single-stage architecture as well as label space representation described in Section \ref{sec:example}, we were able to simultaneously train the neural network to detect objects in 2D and 3D space.  

As mentioned in Section \ref{sec:nn}, the training of the neural network has been framed as a multitask-learning problem. The loss function consists of two parts, 2D and 3D loss terms. The loss function for 2D properties is adapted from YOLO paper \cite{Redmon2018}. For the 3D loss term, the loss has been lifted to 3D instead of calculating the loss for certain individual loss terms (e.g., 2D projection of cuboid center point, orientation). The 3D loss is calculated by reconstructing the bounding cuboid in 3D and then calculating the L2 loss of predicted and ground truth corner points of the cuboid. In addition, the method described in \cite{Simonelli2019} has been utilized to disentangle loss terms. As mentioned before, a masking solution has been utilized to avoid penalizing the network when predicting 3D properties to semi-pseudo-labels that do not have 3D annotations. The final loss is the sum of 2D and 3D loss. 

\section{Experiments}
\label{sec:eval}
We have conducted experiments with the neural network described in Section \ref{sec:example} on a publicly available dataset as well as on internal data. The main goal of the experiments was not to compete with state-of-the-art solutions rather validate the viability of the proposed semi-pseudo-labeling and 3D augmentation methods. Therefore, the baseline is a model trained using neither semi-pseudo-labeling nor 3D augmentations.

\subsection{Argoverse}
\label{sec:argo}
Argoverse \cite{Chang2019} is a collection of two datasets designed to facilitate autonomous vehicle machine learning tasks. The collected dataset consists of 360-degree camera images and long-range LiDAR point clouds recorded in an urban environment. Since the perception range of the LiDAR used for ground truth generation is 200 meters, Argoverse dataset is suitable to validate our methods for enabling long-range camera-only detections. However, LiDAR point cloud itself was not used as an input for the model, only camera frames and corresponding 3D annotation were shown to the neural network. In order to enable semi-pseudo-labeling, the two-dimensional projection of 3D annotations were calculated and an FCOS \cite{Tian2019} model was run on the Argoverse images to obtain 2D detection of unannotated objects. Finally, the deduplication algorithm described in Section \ref{sec:example} has been executed for avoiding multiple containment of objects within the dataset.

\begin{figure*}
  \centering
  \begin{subfigure}{0.48\linewidth}
    \includegraphics[width=0.85\linewidth]{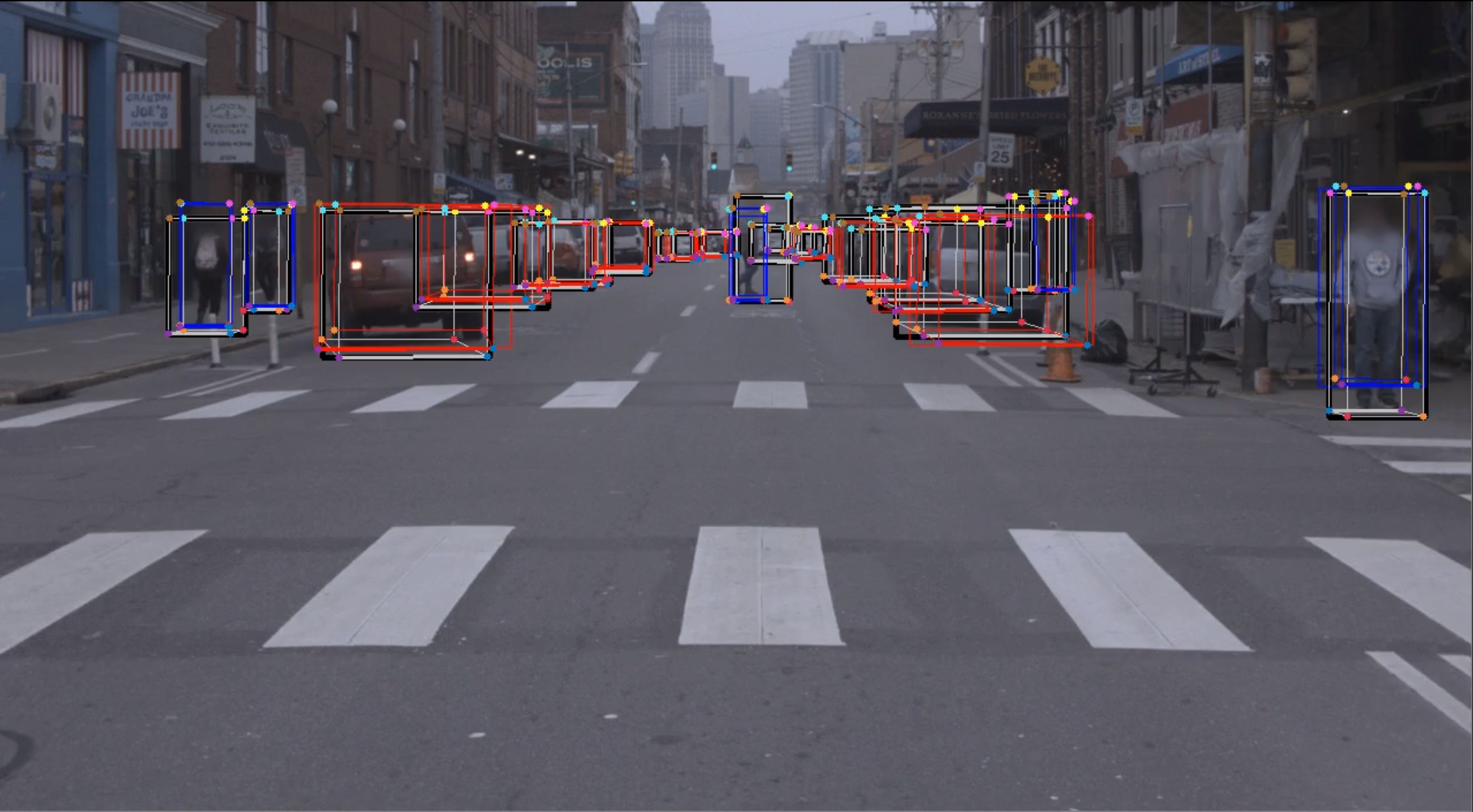}
    \label{fig:av1}
  \end{subfigure}
  \hfill
  \begin{subfigure}{0.48\linewidth}
    \includegraphics[width=0.85\linewidth]{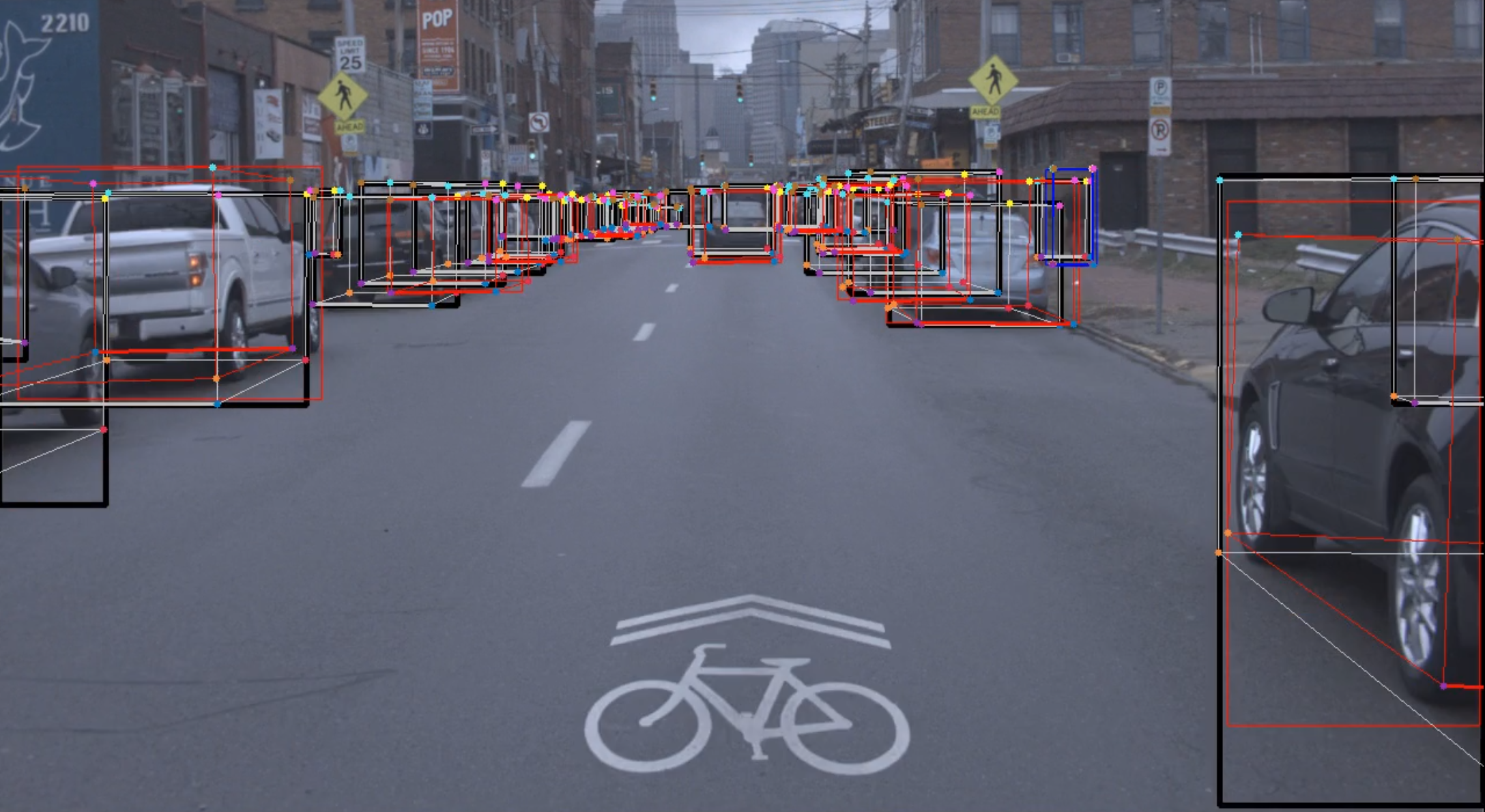}
    \label{fig:av2}
  \end{subfigure}
  \caption{Qualitative results, depicting detections on Argoverse dataset. White cuboids: 3D annotation, black rectangles: 2D projection of 3D annotations, cuboids and rectangles with other colors: 2D and 3D detections of the model.}
  \label{fig:qual}
\end{figure*}

The performance of the models have been measured both in image and model space. Image-space detections indicate the projected 2D bounding boxes of 3D objects while model space is represented as Bird's-Eye-View (BEV) and is used for measuring detection quality in 3D space. Table \ref{tab:argoverse_car} shows the difference between the performance of the baseline model and a model trained using our novel training method for the category 'Car'. A solid improvement in both 2D and BEV metrics can be observed. Since category 'Car' is highly over-represented in the training data with small number of image-visible but not annotated ground truth (these objects are located in the far-region), the performance improvement is not as visible as in the case of other, less frequent object categories. The reason for the low values corresponding to BEV metrics is that the ground truth-prediction assignment happens using the Hungarian algorithm \cite{Kuhn1955} based on the intersection over union metric (IoU threshold is set to $0.5$). Since the longitudinal error of predictions is increasing as the detection distance increases, the bounding box association in BEV space might fail even though the image space detection and association was successful. Figure \ref{fig:qual} depicts some example detection on the Argoverse dataset where distant objects are successfully detected.

The image-space and BEV metrics are shown in Table \ref{tab:argoverse_car} for category 'Large vehicle'. The effect of semi-pseudo-labeling and 3D augmentations can even more be observed than in the case of the 'Car' category. The performance of the baseline model in BEV-space is barely measurable due to the very strict ground truth-prediction assignment rules and heavy class imbalance. This explains the difference between the baseline and the proposed method on BEV Precision metric too. The baseline provides only a few detections in far range with high precision. Our proposed method is able to detect in far-range too (c.f. the difference between 2D and BEV recall of baseline and proposed method), but due to the strict association rules, the BEV precision is low.  Overall, the model trained with our method has significantly better performance in BEV-space as well as in image space. Table \ref{tab:argoverse_ped} shows a similar effect in the case of the 'Pedestrian' category. The BEV metrics are omitted since the top-view IoU-based bounding box assignment violates the association rules due to the small object size. 

\begin{table*}[t]
\caption{Results on Argoverse dataset for 'Car' and 'Large vehicle' objects.}
\label{tab:argoverse_car}
  \centering
  \begin{tabular}{@{}lclclclclclc@{}}
    \hline
    CAR & 2D AUC & 2D Recall & 2D Precision & BEV AUC & BEV Recall & BEV Precision \\
    \hline
    Baseline & 0.5584 & 0.5681 & 0.7773 & 0.1035 & 0.1602 & 0.4161 \\
    Ours & \textbf{0.6307} & \textbf{0.6061} & \textbf{0.7920} & \textbf{0.1180} & \textbf{0.1800} & \textbf{0.4247} \\
    \hline
    \hline
    LARGE VEHICLE & 2D AUC & 2D Recall & 2D Precision & BEV AUC & BEV Recall & BEV Precision \\
    \hline
    Baseline & 0.1412 & 0.1421 & 0.4406 & 0.0338 & 0.0214 & \textbf{0.8636} \\
    Ours & \textbf{0.3921} & \textbf{0.4304} & \textbf{0.5481} & \textbf{0.0767} & \textbf{0.1566} & 0.3798 \\
    \hline
  \end{tabular}
\end{table*}

\begin{table*}[t]
\caption{Results on Argoverse dataset for 'Pedestrian' objects.}
\label{tab:argoverse_ped}
  \centering
  \begin{tabular}{@{}lclclclclclc@{}}
    \hline
    PEDESTRIAN & 2D AUC & 2D Recall & 2D Precision \\
    \hline
    Baseline & 0.1397 & 0.1510 & 0.4985 \\
    Ours & \textbf{0.2745} & \textbf{0.2884} & \textbf{0.5353}  \\
    \hline
  \end{tabular}
\end{table*}

\subsection{In-house Highway Dataset}
\begin{figure*}
  \centering
  \begin{subfigure}{0.48\linewidth}
    \includegraphics[width=0.85\linewidth]{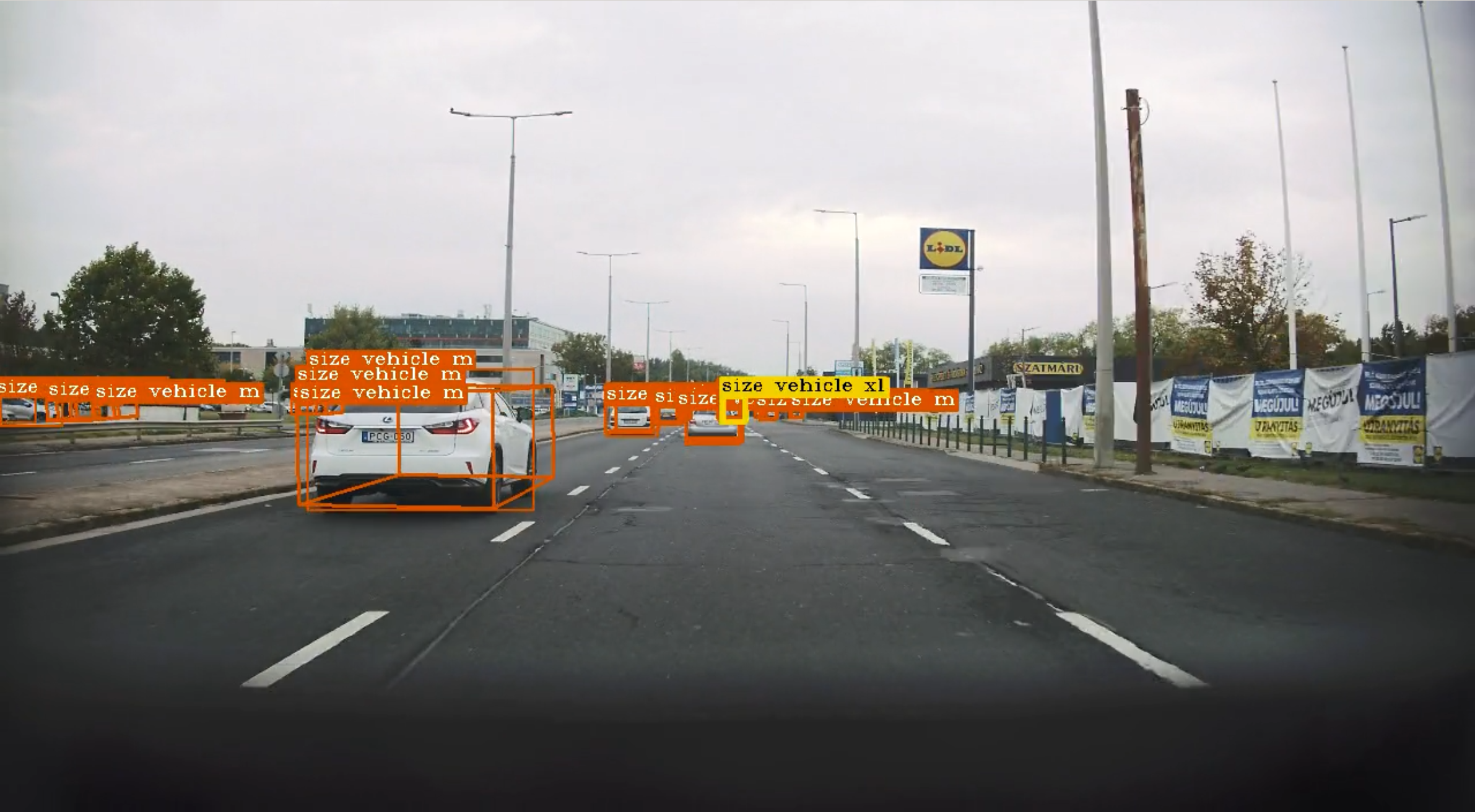}
    \caption{Detections of the model trained using our proposed methods.}
    \label{fig:dets}
  \end{subfigure}
  \hfill
  \begin{subfigure}{0.48\linewidth}
    \includegraphics[width=0.85\linewidth]{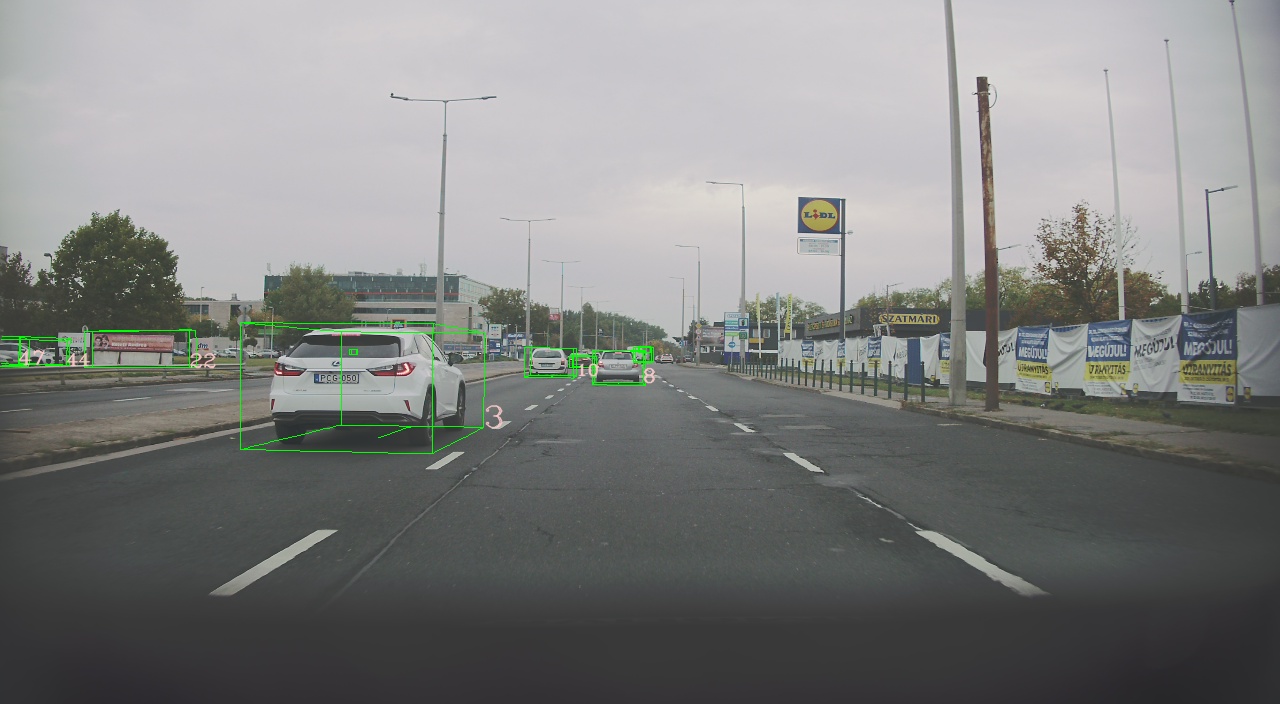}
    \caption{Ground truth with missing distant objects.}
    \label{fig:gts}
  \end{subfigure}
  \caption{The distant objects missing from the ground truth are detected by the model.}
  \label{fig:det_vs_gt}
\end{figure*}

Since the operational domain of Argoverse dataset is urban environment and the validation of our method in highway environment is also a requirement, we have performed an in-house data collection method and created 3D bounding box annotations using semi-automated methods. The sensor setup used for the recordings consisted of four cameras and a LiDAR with a 120-meters perception range both in front and back directions. Figure \ref{fig:det_vs_gt} shows the projected cuboids of a semi-automated annotation sample. It can be observed that distant objects (rarely objects in near/middle-distance region too) are not annotated due to the lack of LiDAR reflections. As a consequence of the limited perception range of the LiDAR, a manual annotation step was needed for creating a validation set. In this way, distant objects (up to 200 meters) and objects without sufficient LiDAR reflections can also be labeled and a consistent validation set can be created. The collected dataset consists of Car, Van, Truck, and Motorcycle categories.
\label{sec:inhouse}

\begin{figure}
  \centering
  \begin{subfigure}{0.45\linewidth}
    \includegraphics[width=0.8\linewidth]{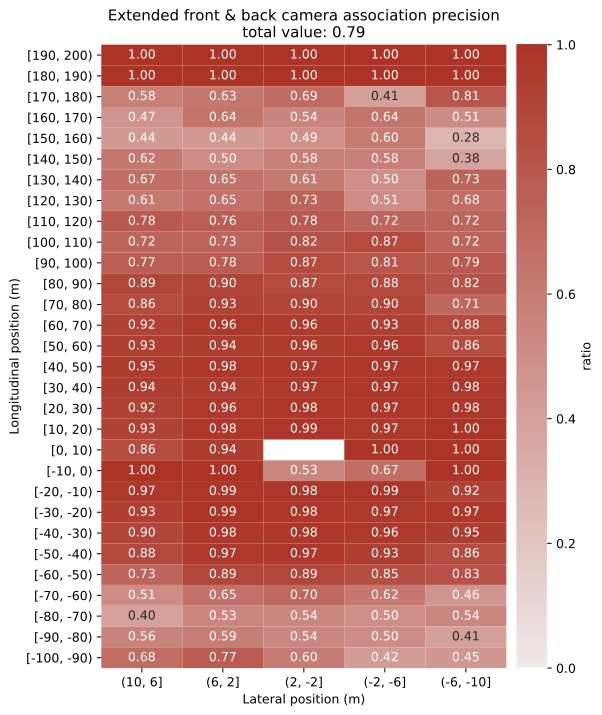}
    \label{fig:prec}
  \end{subfigure}
  \hfill
  \begin{subfigure}{0.45\linewidth}
    \includegraphics[width=0.8\linewidth]{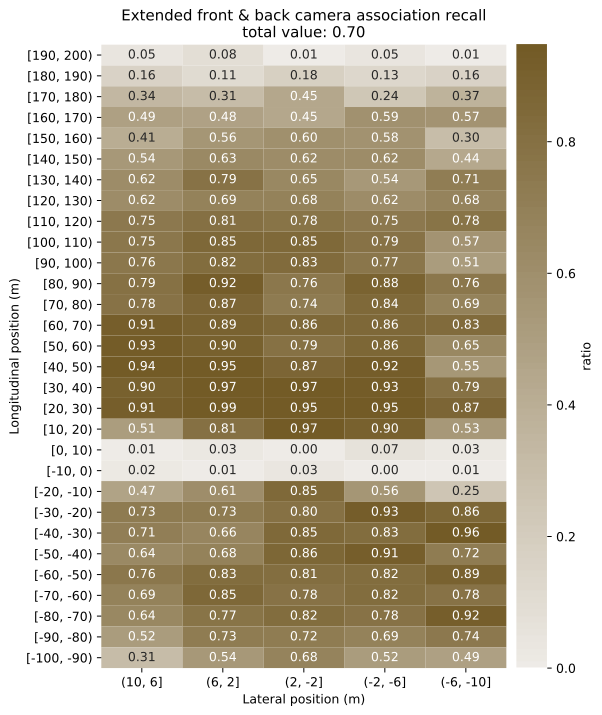}
    \label{fig:rec}
  \end{subfigure}
  \caption{Precision and recall metrics displayed as a heatmap around the ego car.}
  \label{fig:benchmark}
\end{figure}
The model was trained on the semi-automatically annotated data and validated on the manually annotated validation set. Figure \ref{fig:benchmark} shows benchmark results (namely precision and recall metrics) of the neural network trained with our method in a class-agnostic manner. The heatmaps visualize the top-view world space around the ego car where the world space is split into 4 meters by 10 meters cells. The blank cell in the left heatmap indicates the ego car position and can be seen as the origin of the heatmap. The total values on the figure are the average over the heatmap. A prediction is associated with a ground truth if the distance between them is less than 10 meters. The forward detection range is 200 meters while the backward range is 100 meters. It can be observed that the model is able to detect up to 200 meters in forward direction even though the training data did not contain any annotated objects over 120 meters. The low recall value in a near range (-10m, 10m) can be explained by the fact that the model was trained only with front and back camera frames, and objects in this detection area might not be covered by the field-of-view of the camera sensors. The high precision in (180m, 200m) can be attributed to the fact that the model produces only a few detections in the very far range with high confidence (i.e. the model does not produce a large number of false-positive detections in exchange for the larger number of false-negative detections).

Three-dimensional zoom augmentation without semi-pseudo-labeling could not have been able to perform similarly due to the issues described in Section \ref{sec:vanilla}. However, a limitation can be observed since the detection ability significantly drops over 150 meters, as the heatmap of recall metric shows in Fig. \ref{fig:benchmark}.

\section{Conclusion}
\label{sec:conclusion}
In this paper, we have introduced a novel training method for facilitating training neural networks used in the autonomous driving domain. The 3D augmentations have the advantageous effect that it is possible to accurately detect objects that are not part of the training distribution (i.e. detect distant objects without ground truth labels). It is true that semi-pseudo-labelling alone can be enough for the detections, however, the 3D properties, especially depth estimation would be suboptimal due to the fact that neural  networks cannot extrapolate properly outside of the training distribution. Since our main interest was to validate the viability of the proposed method, we used a simple model for the experiments. A future research direction could be to integrate semi-pseudo-labeling and 3D zoom augmentation into state-of-the-art models and conduct experiments in order to examine the effects of our method.
%
%
%
\bibliographystyle{splncs04}
\bibliography{refs}

\end{document}